# High-dimensional Bayesian Optimization Algorithm with Recurrent Neural Network for Disease Control Models in Time Series

Yuyang Chen, Kaiming Bi, Chih-Hang J. Wu, David Ben-Arieh, Ashesh Sinha

**Abstract**

Bayesian Optimization algorithm has become a well-received approach for nonlinear global optimization problems and many machine learning applications. Over the past few years, improvements and enhancements have been brought forward and they have shown some promising results in solving the complex dynamic problems, systems of ordinary differential equations where the objective functions are computationally expensive to evaluate. Besides, the straightforward implementation of Bayesian Optimization algorithm performs well merely for optimization problems with 10-20 dimensions. Study presented in this paper proposes a new high-dimensional Bayesian Optimization algorithm combining Recurrent neural network (RNN-BO algorithm), which is expected to predict the optimal solution for the global optimization problems with high-dimensional or time-series decision models. The proposed RNN-BO algorithm can solve the optimal control problems in the lower dimension space, and then learn from the historical data using the recurrent neural network to learn the historical optimal solution data and predict the optimal control strategy for any new initial system value setting (initial parameter values or initial system state values). In addition, accurately and quickly providing the optimal control strategy is essential to effectively and efficiently control the epidemic's spread while minimizing the associated financial costs. Therefore, to verify the effectiveness of the proposed algorithm, computational experiments are carried out on a deterministic SEIR epidemic model and a stochastic SIS optimal control model. Finally, we also discuss the impacts of different number of the RNN layers and training epochs on the trade-off between solution quality and related computational efforts.

## 1. Introduction

Providing the optimal control strategies for epidemic models has increasingly attracted attentions from both research and health organizations or agencies. During the epidemic, the health organizations or agencies may take a series of control strategies (epidemic prevention or intervention measures) for mitigating the local outbreak, e.g., vaccination, quarantine, disinfection, or regional closures. All these control measures could be associated with certain financial costs, directly or indirectly. If health organizations or agencies do not control the epidemic, it may also cause inevitable economic consequences, such as workforce losses due to outbreaks, increased community healthcare costs, local business downturns, and declined related travels. Thus, the optimal control strategy of epidemic should balance the corresponding financial cost of control and the epidemic progression. During recent years a number of studies on the optimal control to control the spread of epidemic and relieve healthcare financial burden have been carried out [1-4]. Therefore, it is important for public health purposes to figure out the optimal control policy for the trade-off of strategies effectiveness and cost efficacy [1].

Although many works on studying the optimal control strategy for different epidemic diseases have been made, their results only provided the optimal control policy for the specific regions (city/state/country) or virus type of epidemic. Those optimal control strategies may no longer be effective or optimal if the epidemic outbreaks in different regions or the epidemic viruses mutate. For instance, when the influenza viruses mutate into three different types, the control treatment for a specific type of virus cannot work for the infective individuals infected by another type of virus or virus variation [5]. This means that researchers need to build a new model to study the relationship of optimal control and epidemic progression caused by different virus. At the end of 2019, the COVID-19 emerged in Wuhan, China and rapidly spread to the rest of the world. Due to the different transmission dynamics and economic situation of different regions of the world, the optimal control policies and the related control intensity usually are different [6]. Recently, due to the variant of COVID-19 virus, the unprecedented increase happens in several countries. The existing vaccines may lack the efficacy on controlling the variant virus, which means the government officials need to re-plan the control policy regarding the new situation of the epidemic [7]. The above challenge offers guidance for authors to focus on the algorithm development of optimal control learning and prediction. Since the initial epidemic system value setting is associated with the epidemic regions or virus types. The target algorithm is expected to have ability to learn the historical epidemic data (including initial system state/parameter data and corresponding optimal control strategy data), and then quickly predict the new optimal control solution to respond to new epidemic region or virus variation.

Our main purpose is to develop such a general learning and prediction algorithm, which can predict the optimal control solution only based on the known historical epidemic data, even though this data is from different epidemic regions or different types of viruses of same epidemic. For example, for some poor regions or countries, they may have no enough financial supports to collect local epidemic data. However, the data of other regions or countries under the same epidemic is available to access, it will be helpful and meaningful if the poor regions or countries can leverage those available data to generate the effective control policy for themselves though the data is from different places. Besides this, it's necessary to develop the framework according to some characteristics of the epidemic control model, which can guarantee the effectiveness and efficiency of the framework. To our knowledge, the epidemic control model with hundreds of thousands of time epochs is considered as high-dimensional and time-series [8]. If we consider each time epoch as one dimension of the model, hence the epidemic model is high-dimensional. Also, time-series means the values of the state variables and control variables of the model at current time will affect subsequent state variable values and control variable values. Therefore, considering these characteristics, we intent to develop a new algorithm combining a high-dimensional global optimization algorithm and Recurrent Neural Network (RNN) algorithm.

Conventional global optimization techniques, such as particle swarm optimization algorithm, genetic algorithm, simulated annealing algorithm, stochastic gradient descent, etc., are suitable for solving the low-dimensional systems with the nature of time-independent or dimensions-independent. They may not be enough effective to solve high-dimensional optimization systems. Bayesian Optimization (BO) is popular and powerful global optimization approach, it is also computationally challenging to handle high-dimensional systems [9]. However, BO is a promising learning-based method to dig out more hidden or posterior information from historical data. Some improved BO algorithms are proposed for handling the high-dimensional global optimization problems. For example, Moriconi *et. al.* proposed a high-dimensional BO algorithm by learning a nonlinear feature mapping to reduce the inputs' dimension, this improvement allows the algorithm to easily optimize the acquisition function in low-dimensional space [10]. Zhang *et. al.* introduced a sliced inverse regression method to BO to learn the intrinsic low-dimensional structure of the objective function in high-dimensional space, which can automatically study the intrinsic structure of objective function during the optimization process [11]. Li *et. al.* developed a new method for high-dimensional BO by using dropout strategy to reduce dimensions and optimize a subset of variables, and then provide three "fill-in" strategies to guide how to fill-in the left-out dimensions at each acquisition function optimization [12]. However, these high-dimensional BO algorithms need to reconstruct the variables from low-dimensional space back into its original dimension space at each iteration, and then calculate the corresponding objective function value in high-dimensional space. Although those new BO algorithms overcome the high-dimensional difficulty, they didn't realize computational efficiency. Also, those BO algorithms are not suitable to solve the time-series optimization system. Therefore, we attempt to develop the target algorithm combining an BO algorithm with RNN algorithm. The BO algorithm used in the target algorithm is like the improved BO algorithm proposed in our previous work [8].

The target algorithm is in short denoted as RNN-BO algorithm. By combining RNN algorithm, the proposed algorithm keeps the advantage of RNN to learn the optimal control data on past epidemics (epidemics from same/different regions), then predict the optimal control strategy toward future outbreaks happened in different regions or happened due to virus variation. Our extensive computational experiments have shown that the RNN-BO algorithm can effectively overcome some of the shortcomings in the existing high-dimensional BO algorithms and epidemic optimal control optimization algorithms. In this paper, we use the time-dependent deterministic SEIR and stochastic SIS epidemic control models to illustrate the benefits and advantages of the RNN-BO algorithm. The main contributions of this paper are summarized below:

- Propose a novel RNN-BO algorithm that is effective and computationally efficient for high-dimensional global optimization problems with time-series epidemic model.
- The RNN-BO algorithm is capable to learn the relationship between the optimal control solution and initial system value setting of complex epidemic model. Then construct a prediction model based on historical data,

- which can quickly and accurately predict the corresponding optimal control strategy once given any new initial system value setting (initial system state values or system parameter values).
- The RNN-BO algorithm reduces the high-dimensional variables into low-dimensional space when solves optimal control solution using BO algorithm, it does not require to reconstruct the control variables back to high-dimensional space at each iteration.
- The RNN-BO algorithm takes advantage of historical epidemic data from different regions or virus variation, which can continually learn and modify the prediction model so that it can offer effective and accurate optimal control strategy even there is less knowledge about new outbreak of same epidemic.

The rest of the paper is structured as follows. Section 2 introduces the epidemic control optimization systems as the application under this study. Section 3 provides the background and presents the RNN-BO algorithm in detail. Section 4 demonstrates the effectiveness of the RNN-BO algorithm and makes comparison with the standard Bayesian Optimization algorithm and a high-dimensional Bayesian Optimization algorithm through numerical simulation experiments. Finally, conclusions and future works are drawn in Section 5.

## 2. Problem Formulation

The model we attempt to research is high-dimensional time-series epidemic control model. In this paper, we plan to research the RNN-BO algorithm on two different high-dimensional time-series epidemic control models: deterministic SEIR control model and stochastic SIS control model. These two control models are developed based on the standard deterministic SEIR [13] and stochastic SIS model [14], respectively. Two original standard epidemic models didn't consider the control variables, which only can be used to study the natural progression of the epidemic without any control strategy (epidemic prevention or intervention). However, in real world, the health organizations or agencies usually take a series of control strategies for mitigating the local outbreak.

The control strategy can affect contact rate or infection rate between individuals and finally affect the progression of the epidemic. For example, during COVID-19 epidemic, CDC noticed that control measures like facemasks, cloth mask or respirator can prevent the spread of respiratory secretions. Those control measures provide different levels of protection for people against exposure to infectious droplets and particles produced by infected people [15]. Also, encouraging and guaranteeing the safe social distance is a control strategy to decrease the contact and infection rate [16]. The control variables can represent the level/degree/intensity of restrictions on activities, mask wearing, quarantines or medicine care, it also can represent the vaccination coverage rates [16-18]. These facts indicate that the contact rate or infected rate can be controlled through practical control approaches, the control variable of the epidemic models in this paper has practical meaning, and is truly controllable. Thus, we consider the control variables into the standard SEIR and SIS epidemic model, and solve the optimal control strategy that minimizes the overall financial cost associated with control strategies and controls the spread of the epidemic.

The optimization system with the deterministic SEIR control model developed based on standard SEIR model [13] is formulated as follows:

$$\text{Min } V = \int_{t_1}^{t_f} C_1 I(t) + C_2 f(u_1, u_2, t) \tag{1}$$

$$s.\,t. \quad \frac{dS(t)}{dt} = \tau - (1 - u_1(t))\beta S(t)I(t) - \tau S(t) \tag{2}$$

$$\frac{dE(t)}{dt} = (1 - u_1(t))\beta S(t)I(t) - (\tau + \alpha)E(t) \tag{3}$$

$$\frac{dI(t)}{dt} = \alpha E(t) - (\tau + \gamma)I(t) - u_2(t)I(t) \tag{4}$$

$$\frac{dR(t)}{dt} = \gamma I(t) - \tau R(t) + u_2(t)I(t) \tag{5}$$

$$S(t) + E(t) + I(t) + R(t) = 1 \tag{6}$$

where $t_1$ is the start time of control, $t_f$ is the end time of control. $V$ indicates the overall cost due to control measures and the cost due to infected population if the health organizations or agencies don't take any control measure during time period $[t_1, t_f]$. The parameters $C_1$ and $C_2$ in Eq. (1) represent the financial cost of system without control and with control, respectively.

$f(u_1, u_2, t)$ in Eq. (1) is the cost function due to the control strategy. Most existing studies consider the cost function associated with the control strategy as convex [19, 20]. However, in the real world, the cost function is possible non-convex [21]. To better verify the effectiveness and efficiency of the proposed RNN-BO algorithm, the cost function $f(u_1, u_2, t)$ will be considered as non-convex in this paper.

$S(t), E(t), I(t), R(t)$ in Eqns. (1) – (6) are the system state variables, they represent the fraction of susceptible, exposed, infected, and recovery population at time $t$, respectively. $S$ represents the individuals who might be infected the disease; $E$ represents the individuals who have been infected but are not infectious, they are not capable to transmit the disease; $I$ represents the individuals who have been infected and are able to transmit the disease; $R$ represents the individuals who have become immune.

Parameter $\tau$ in Eqns. (2) – (5) represents the natural birth rate, we assume the natural death rate is equal to the natural birth rate in this paper. Parameter $\beta$ in Eqns. (2) - (3) represents the natural contact rate between $S$ and $I$ when there is no any control strategy like quarantine or activity restriction. Parameter $\alpha$ in Eqns. (3) – (4) represents transfer rate from state $E$ to $I$. Parameter $\gamma$ in Eqns. (4) - (5) represents natural recovery rate from state $I$ to $R$ when there is no any control strategy like medicine treatment or hospitalization.

$u_1$ and $u_2$ are the system decision variables (control variables), their values represent the level/degree/intensity of the corresponding control measures. $u_1$ in Eqns. (2) – (3) represents prevention control strategy that can slow down the probability of $S$ being infected by $I$, such as vaccination, quarantine, activity restriction, social distance restriction, etc. $u_2$ in Eqns. (4) – (5) represents intervention control strategy that can speed up the population's recovery from state $I$ to state $R$, such as medicine treatment, hospitalization, advanced medical facilities and equipment, etc.

Assume each control strategy contains $(\mathcal{D} = t_f - t_1)$ time epochs. If each time epoch is considered as one time dimension of the system, it means the control variable is $\mathcal{D}$ dimensions. Define the control variables as $u_1 = \{u_1(t_1), ..., u_1(t_f)\}$ and $u_2 = \{u_2(t_1), ..., u_2(t_f)\}$, where $u_1(t), u_2(t) \in [u_l, u_u]$ ($t_1 \leq t \leq t_f$), $u_l$ and $u_u$ represent the lower and upper bound of control variable, respectively. $u_1(t)$ and $u_2(t)$ mean the level/degree of the prevention and intervention control strategy at time $t$, respectively. Therefore, the lower and upper bound of control variables are assumed as $u_l = 0$, and $u_u = 1$.

Next, the stochastic SIS control model studied in this paper only contains two states: susceptible $S$ and infected $I$. In addition, in real world the natural contact rate is possible uncertain, which may be affected by some stochastic environment factors like seasonal variations, climate change, air humidity. Hence, the stochastic SIS control model with a stochastic contact rate is formulated based on standard SIS model [14]:

$$s.\ t.\quad \frac{dS(t)}{dt} = \tau - (1 - u_1(t))\beta S(t)I(t) + \gamma I(t) - \tau S(t) + u_2(t)I(t) - \sigma S(t)I(t)dB(t)/dt \qquad (7)$$

$$\frac{dI(t)}{dt} = (1 - u_1(t))\beta S(t)I(t) - (\tau + \gamma)I(t) - u_2(t)I(t) + \sigma S(t)I(t)dB(t)/dt \qquad (8)$$

$$S(t) + I(t) = 1 \qquad (9)$$

where $B(t)$ is a standard Brownian motion, we use it to describe the uncertainty of the contact rate in the stochastic SIS control model. Eq. (7) means that the stochastic contact rate is normally distributed with mean $\beta dt$ and variance $\sigma^2 dt$, we refer readers to paper [14] to get more exact details and definitions of $B(t)$ and $\sigma$.

In this paper, we use a standard Brownian motion to describe the uncertainty of the contact rate in the stochastic SIS control model. Therefore, the time-series optimal control problem with complex and high-dimensional stochastic SIS epidemic model researched in this paper can be formulated as follows:

$$\text{Min } V = \int_{t_1}^{t_f} C_1 I(t) + C_2 f(u_1, u_2, t) \qquad (10)$$

$$s.\ t.\quad dS(t) = \big(\tau - (1 - u_1(t))\beta S(t)I(t) + \gamma I(t) - \tau S(t) + u_2(t)I(t)\big)dt - \sigma S(t)I(t)dB(t) \qquad (11)$$

$$dI(t) = \big((1 - u_1(t))\beta S(t)I(t) - (\tau + \gamma)I(t) - u_2(t)I(t)\big)dt + \sigma S(t)I(t)dB(t) \qquad (12)$$

$$S(t) + I(t) = 1 \qquad (13)$$

where $B(t)$ denotes the standard Brownian motion with the intensity of noise $\sigma$. The term $(1 - u_1(t))\beta$ and $u_2(t)I(t)$ have the similar meaning as shown in SEIR model.

# 3. High-dimensional Bayesian Optimization Algorithm with RNN

In this section, we develop a new high-dimensional Bayesian Optimization algorithm (RNN-BO algorithm) by combining an improved BO algorithm and RNN for time-series epidemic control models. The BO algorithm used in RNN-BO algorithm is similar to the improved BO algorithm proposed in our previous work [8]. Herein, we just briefly introduce it, we refer readers to paper [8] to get more details of the BO algorithm. The RNN-BO algorithm is capable of predicting a time-series optimal control strategy quickly that can minimize the cost function $V$ and effectively control the disease spread once given any new initial epidemic system value setting. The RNN-BO algorithm includes two part: BO part and RNN part. BO part is mainly to generate enough historical data for further RNN part use. In BO part, we vary the initial system value setting (change the system parameter values or initial system state values), then solve the corresponding final optimal control strategy using the BO algorithm. Store each initial system value setting and corresponding optimal control strategy as one historical data pair. We can obtain many data pairs in the BO part by changing different initial system state values or parameter values. RNN part is to learn all historical data pair (initial system value setting as input and corresponding optimal control strategy as output), then generate a prediction model. This prediction model can be used to predict the optimal control solution once given any new input. In this section, the BO part is briefly introduced from section 3.1 to 3.5. The RNN part is introduced in section 3.6.

## 3.1 Time-dimensions reduction

Unlike the standard Bayesian optimization, the RNN-BO algorithm attempts to solve the optimal control strategy in a low-dimensional space. There are two purposes for making time-dimensions reduction of the control strategy variable. One is to find the optimal solution quickly and accurately, two is to generate data sequences with the nature of time-series for further RNN use. For the control strategy variable with full $\mathcal{D}$ dimensions, we select $d$ dimensions ($d < \mathcal{D}$) of the control variable at each iteration.

## 3.2 Gaussian Process

The surrogate model used in the RNN-BO algorithm is the Gaussian process model. The Gaussian process is used to find the prior belief based on historical data and dig the posterior information, which it's better to evaluate the complex nonconvex objective function and find the optimal solution. For a Gaussian process, we assume that for any control strategy samples $\{\ldots, u^i, \ldots\} \in [0,1]^{t_f - t_1}$, we do the time-dimensions reduction and generate $d$-dimensional $u^i = \{u(t_1), \ldots, u(t_d)\} \in [0,1]^d$ for each control sample, the corresponding objective function value set $[\ldots, V(u^i), \ldots]^T$ for all samples set $\{\ldots, u^i, \ldots\} \in [0,1]^d$ follows the multivariate Gaussian distribution:

$$[\ldots, V(u^i), \ldots]^T \sim \mathcal{GP}(M, K) \tag{14}$$

where $M$ is a mean vector $[\ldots, m(u^i), \ldots]^T$ and $K$ is a covariance matrix as below:

$$K = \begin{bmatrix} k(u^1, u^1) & \cdots & k(u^1, u^i) \\ \vdots & \ddots & \vdots \\ k(u^i, u^1) & \cdots & k(u^i, u^i) \\ k(u^{i+1}, u^1) & \cdots & k(u^{i+1}, u^{i+1}) \\ \vdots & \ddots & \vdots \end{bmatrix} \tag{15}$$

$m(u^i)$ is mean function that is usually defined as a linear function or zero [22]. $k(u^i, u^j)$ is called covariance function or kernel function of two control strategies $u^i$ and $u^j$. There are many different kernel function choices, such as radial basis function (RBF), Matern 52, Linear, Exponential, etc. In the RNN-BO algorithm, we use a common choice Matern 52 as the kernel function [23], it is formulated as:

$$k(u^i, u^j) = (1 + \sqrt{5} * \frac{|u^i - u^j|}{l} + \frac{5}{3} * \frac{|u^i - u^j|^2}{l^2}) \exp\left(-\sqrt{5} * \frac{|u^i - u^j|}{l}\right) \tag{16}$$

where $l$ is the length-scale hyperparameter. Different kernel functions and the value of $l$ may lead to different global optimization performances, which can be tested to pick the better choice through implementing numerical experiments [21]. Since it's not the main contribution of this paper, we will not provide more detail here.

The mean vector $M$ and covariance matrix $K$ can be viewed as the prior belief. From the Gaussian process model, for any new control strategy $u^{new} \in [0,1]^d$, the objective function value $V(u^{new})$ at the new point $u^{new}$ will follows the distribution:

$$V(u^{new})|D \sim \mathcal{GP}(M^{new}, K^{new}) \tag{17}$$

where $D$ is the dataset storing the historical data. $M^{new}$ and $K^{new}$ represent the posterior mean and posterior covariance, respectively. They can be expressed as:

$$D = \{\ldots(u^i, V(u^i)), \ldots\} \tag{18}$$

$$M^{new} = \mu(V(u^{new})|D) = m(u^{new}) + K'K^{-1}(V - M) \tag{19}$$

$$K^{new} = \sigma(V(u^{new})|D) = K'' - K'K^{-1}K'^T \tag{20}$$

where:

$$K' = [k(u^{new}, u^1), \ldots, k(u^{new}, u^i), \ldots] \tag{21}$$

$$K'' = k(u^{new}, u^{new}) \tag{22}$$

$$V = [\ldots, V(u^i), \ldots]^T \tag{23}$$

*3.3 Acquisition Function*

The acquisition function that we used to estimate the original objective function during the optimization process of the RNN-BO algorithm is the lower confidence bound (LCB) function [8]. The goal of using acqusition function is to utilize the posterior information to find a new better sampling point at each iteration, and this new sampling point can balance the purpose of exploration and exploitation. The exploration means that the algorithm tends to sample the next points with highly uncertainty. The exploitation means the algorithm will sample the next points with the lower objective function value in the minimization problems. We know that for any new control strategy, we have the posterior information $\mu(V(u^{new})|D)$ and $\sigma(V(u^{new})|D)$ from Gaussian process. The posterior mean $\mu(V(u^{new})|D)$ can represent the exploitation, the posterior covariance $\sigma(V(u^{new})|D)$ can represent the explotation. The LCB acquisition function can be calculated as:

$$\text{LCB}(u) = \mu(V(u^{new})|D) - k\sigma(V(u^{new})|D) \tag{24}$$

where $k$ is the weight to balance the posterior mean and the covariance. A large value of $k$ indicates that the algorithm places more weight on sampling a new point with high uncertainty, A small value of $k$ indicates that the algorithm places more weight on sampling a new point with a small objective function value. At each iteration, we sample the next control strategy point that minimizes the LCB acquisition function:

$$u^{new} = \arg\min_{u} \text{LCB}(u) \tag{25}$$

*3.4 Sampling Strategy*

When the model is high-dimensional, it is usually impossible to search the entire feasible solution space to solve the optimal control strategy for Eq. (25) at each iteration. Therefore, an effective and efficient sampling strategy is necessary to improve the computational efficiency of the RNN-BO algorithm. In the BO part of the RNN-BO algorithm, we combine the multi-armed bandit and random search to sample new candidates for optimizing the acquisition function.

Multi-armed bandit (MAB) is a class reinforcement learning case of the trade-off between exploration and exploitation [24]. MAB means that we decide to choose one or some bandits from all bandits to play at each iteration. Each bandit is configured with a reward of how the decision-maker will likely earn a reward regarding the decision. In the RNN-BO algorithm , the steps using MAB to sample the new candidates are: (1) Divide the range of control strategy into some small ranges and consider them as bandits; (2) Define the reward of each small range is equal to $n_{MAB}$, and at each small range, sample $n_{MAB}$ candidate points; (3) Calculate the corresponding acquisition function values for those sampling candidates; (4) Find the largest and smallest acquisition function values, and update the reward $n_{MAB}$ of each range; (5) Repeat (2)-(4) for some iterations, and then find the best candidate point $u_{MAB}^{new}$ with the best acquisition function. Here, in step (4), we assume the range that the candidate point with the largest acquisition function value belongs to will earn one reward, the reward at this range will be updated as $n_{MAB} \leftarrow n_{MAB} + 1$, which means that the RNN-BO algorithm will sample $n_{MAB} + 1$ candidate points from this range at the next iteration. The range that the point with the smallest value belongs to will lose one reward, the reward at this range will be updated as $n_{MAB} \leftarrow$ $n_{MAB} - 1$, which means that the algorithm will sample $n_{MAB} - 1$ candidate points from this range at the next iteration. Fig. 1 is a sampling point example of MAB.

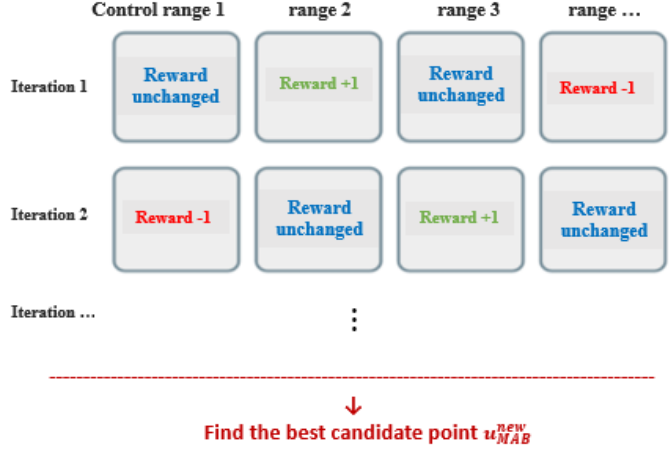

**Fig. 1. A sampling point example of MAB**

Random research is another sampling method in the BO part. We randomly sample $n_{RS}$ candidate points with lower bound $u_l$ and upper bound $u_u$. Then calculate the corresponding acquisition function values for all candidates, and pick the best one with the lowest acquisition function value as the optimal candidate $u_{RS}^{new}$ generated by random search. At each iteration, we compare the corresponding acquisition function values of candidate $u_{MAB}^{new}$ and $u_{RS}^{new}$, choose the better one as the optimal new sampling point:

$$u^{new} = \begin{cases} u_{MAB}^{new}, & if \ \text{LCB}(u_{MAB}^{new}) < \text{LCB}(u_{RS}^{new}) \\ u_{RS}^{new}, & if \ \text{LCB}(u_{MAB}^{new}) > \text{LCB}(u_{RS}^{new}) \end{cases} \quad (26)$$

After some iterations, we can get the optimal solution $u^*$ by comparing all $u^{new}$.

*3.5 Local Search*
To increase the final solution's accuracy, we add a local search after the acquisition function optimization. Since Adam method can faster converge to a local minimum with better quality [25]. Therefore, in the BO part of the RNN-BO algorithm, a local search based on Adam gradient descent is implemented starting from the optimal solution $u^*$. The final $d$-dimensional time-series optimal control solution is obtained after the local search.

*3.6 Bayesian Optimization with Recurrent Neural Network*
The proposed RNN-BO algorithm is expected to quickly and accurately predict the optimal control strategy when given any new initial system value setting. Therefore, this paper applies RNN to learn the historical data pairs obtained from the BO part, and find the relationship between the optimal control strategy and system initial system value setting. This section will describe how to utilize RNN to learn the historical data pairs and generate a prediction model.

RNN is a type of artificial neural network widely used to process sequential data or time series data, which is demonstrated to produce state-of-the-art results in various sequence learning problems [26]. Different from traditional neural networks, the inputs and outputs of RNN are dependent on each other. RNN depends on the prior data within the sequence, which utilizes the training data to learn the feature and position information. An application example of RNN is described in detail in [27]. An excellent advantage of RNN is it can take one data or a series of data in time order as input and produce one value or a series of values as output. Therefore, there are many different types of RNN due to various inputs and outputs in length [28]. In the RNN-BO algorithm, we decide to use the one-to-one RNN, which means that the algorithm maps one input vector to one output.

Next, we introduce how to design the input and output data obtained from the BO part for further RNN training use. We use the SEIR model as an example. As shown in Fig. 2, at iteration 1, initialize the system state values $(S_1(t_1), E_1(t_1), I_1(t_1), R_1(t_1))$, determine the values of system parameters, and randomly generate a $d$-dimensional control strategy. Then we can calculate all system state values in $d$ time dimensions through Eqns. (2) – (6). We solve $d$-dimensional optimal control strategy using the BO algorithm described from section 3.1 to 3.5. Then update all state values from time $t_1$ to $t_d$. After iteration 1, we can obtain the data from time $t_1$ to $t_d$ as:

$$\begin{bmatrix} S_1(t_1) & S_1(t_2) & S_1(t_3) & \ldots & S_1(t_{d-1}) & S_1(t_d) \\ E_1(t_1) & E_1(t_2) & E_1(t_3) & \ldots & E_1(t_{d-1}) & E_1(t_d) \\ I_1(t_1) & I_1(t_2) & I_1(t_3) & \ldots & I_1(t_{d-1}) & I_1(t_d) \\ R_1(t_1) & R_1(t_2) & R_1(t_3) & \ldots & R_1(t_{d-1}) & R_1(t_d) \end{bmatrix} \tag{27}$$

$$\{u_1(t_1) \quad u_1(t_2) \quad u_1(t_3) \quad \ldots \quad u_1(t_{d-1}) \quad u_1(t_d)\} \tag{28}$$

Then, we choose the state value $(S_1(t_2), E_1(t_2), I_1(t_2), R_1(t_2))$ at time $t_2$ in Eq. (27) as the initial state values for iteration 2. Under the same parameter values, randomly generate a $d$-dimensional control strategy and calculate the state values, then optimize the control strategy and update the state values (do the same thing as iteration 1). After iteration 2, we can obtain the data from time $t_2$ to $t_{d+1}$ as:

$$\begin{bmatrix} S_1(t_2) & S_2(t_3) & S_2(t_4) & \ldots & S_2(t_d) & S_2(t_{d+1}) \\ E_1(t_2) & E_2(t_3) & E_2(t_4) & \ldots & E_2(t_d) & E_2(t_{d+1}) \\ I_1(t_2) & I_2(t_3) & I_2(t_4) & \ldots & I_2(t_d) & I_2(t_{d+1}) \\ R_1(t_2) & R_2(t_3) & R_2(t_4) & \ldots & R_2(t_d) & R_2(t_{d+1}) \end{bmatrix} \tag{29}$$

$$\{u_2(t_2) \quad u_2(t_3) \quad u_2(t_4) \quad \ldots \quad u_2(t_d) \quad u_2(t_{d+1})\} \tag{30}$$

Then choose the state value $(S_2(t_3), E_2(t_3), I_2(t_3), R_2(t_3))$ at time $t_3$ in Eq. (29) as the initial state values for iteration 3. Do the same things and obtain the data from $t_3$ to $t_{d+2}$, and so on. Stop the algorithm until it obtains the data from time $t_{f-d+1}$ to $t_f$ as:

$$\begin{bmatrix} S_{f-d}(t_{f-d+1}) & S_{f-d+1}(t_{f-d+2}) & S_{f-d+1}(t_{f-d+3}) & \ldots & S_{f-d+1}(t_{f-1}) & S_{f-d+1}(t_f) \\ E_{f-d}(t_{f-d+1}) & E_{f-d+1}(t_{f-d+2}) & E_{f-d+1}(t_{f-d+3}) & \ldots & E_{f-d+1}(t_{f-1}) & E_{f-d+1}(t_f) \\ I_{f-d}(t_{f-d+1}) & I_{f-d+1}(t_{f-d+2}) & I_{f-d+1}(t_{f-d+3}) & \ldots & I_{f-d+1}(t_{f-1}) & I_{f-d+1}(t_f) \\ R_{f-d}(t_{f-d+1}) & R_{f-d+1}(t_{f-d+2}) & R_{f-d+1}(t_{f-d+3}) & \ldots & R_{f-d+1}(t_{f-1}) & R_{f-d+1}(t_f) \end{bmatrix} \tag{31}$$

$$\{u_{f-d+1}(t_{f-d+1}) \quad u_{f-d+1}(t_{f-d+2}) \quad u_{f-d+1}(t_{f-d+3}) \quad \ldots \quad u_{f-d+1}(t_{f-1}) \quad u_{f-d+1}(t_f)\} \tag{32}$$

After iterations, we obtain the data for the specific initial system state value setting $(S_1(t_1), E_1(t_1), I_1(t_1), R_1(t_1))$ and system parameter value setting. By changing the initial system state value or system parameter values and do the same process, then we can obtain many data and consider all those data as historical data.

Now we design the training inputs and outputs using those historical data. Consider the one-to-one RNN in the RNN-BO algorithm, we denote the system value setting $(S_{iteration}(t_i), E_{iteration}(t_i), I_{iteration}(t_i), R_{iteration}(t_i), \beta)$ as one input vector, where $\beta$ is the specific infection rate (system parameter). The correspond output is the control value $u_{iteration}(t_i)$ at time $t_i$. Thus, for a specific initial system value setting $(S_1(t_1), E_1(t_1), I_1(t_1), R_1(t_1), \beta)$, there are $d * (f - d + 1)$ input-output data pairs as shown in Table 1. The RNN-BO algorithm doesn't require that the input has to be all system state variables or all system parameters, such as the input as $(S_{iteration}(t_i), I_{iteration}(t_i), \beta)$. The objective function values or other system parameter values also can be used as inputs. We can adjust different elements as input according to the accuracy of final prediction solution.

If enough data is ready to use by changing different initial system value setting, we apply RNN to learn the data and generate a prediction model named RNN-BO prediction model. For any new initial system state values of the same epidemic, we don't have to implement the BO algorithm to solve the optimal control strategy through several optimization iterations. We only need to use the RNN-BO prediction model to predict the optimal control value at the beginning time, then calculate the state values for the next time using Eqns. (2) – (6). After that, the prediction model will predict the optimal control value at the next time. Repeat the process until we obtain $t_f$-dimensional time-series optimal control strategy. Once the RNN-BO prediction model is ready, the algorithm can easily and accurately predict the time-series optimal control strategy. The computational time of prediction process only takes a few seconds. The excellent computational efficiency and global optimization performance of the RNN-BO algorithm will be demonstrated in later simulation section. The implementation flowchart of the RNN-BO algorithm is shown in Fig. 3.

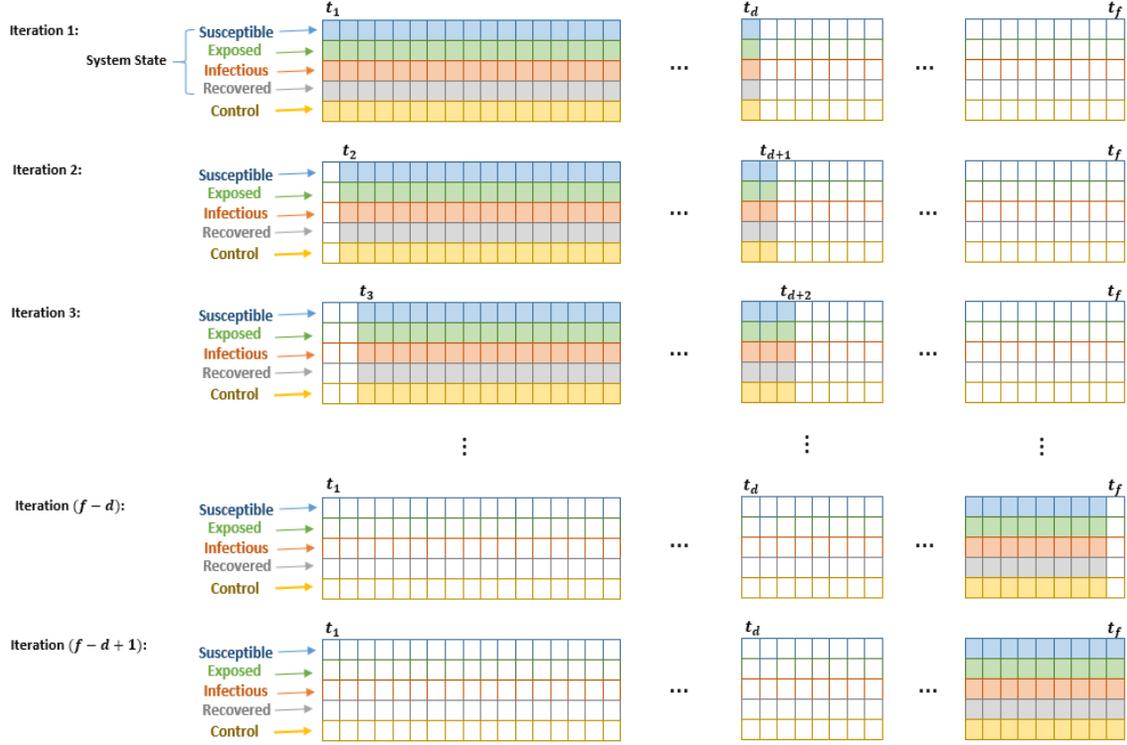

**Fig. 2. Data collection process of the RNN-BO algorithm**

| inputs | outputs |
|---|---|
| $(S_1(t_1), E_1(t_1), I_1(t_1), R_1(t_1), \beta)$ | $u_1(t_1)$ |
| $(S_1(t_2), E_1(t_2), I_1(t_2), R_1(t_2), \beta)$ | $u_1(t_2)$ |
| $\vdots$ | $\vdots$ |
| $(S_1(t_d), E_1(t_d), I_1(t_d), R_1(t_d), \beta)$ | $u_1(t_d)$ |
| $(S_1(t_2), E_1(t_2), I_1(t_2), R_1(t_2), \beta)$ | $u_2(t_2)$ |
| $(S_2(t_3), E_2(t_3), I_2(t_3), R_2(t_3), \beta)$ | $u_2(t_3)$ |
| $\vdots$ | $\vdots$ |
| $(S_2(t_{d+1}), E_2(t_{d+1}), I_2(t_{d+1}), R_2(t_{d+1}), \beta)$ | $u_2(t_{d+1})$ |
| $\vdots$ | $\vdots$ |
| $(S_{f-d}(t_{f-d+1}), E_{f-d}(t_{f-d+1}), I_{f-d}(t_{f-d+1}), R_{f-d}(t_{f-d+1}), \beta)$ | $u_{f-d+1}(t_{f-d+1})$ |
| $\vdots$ | $\vdots$ |
| $(S_{f-d+1}(t_f), E_{f-d+1}(t_f), I_{f-d+1}(t_f), R_{f-d+1}(t_f), \beta)$ | $u_{f-d+1}(t_f)$ |

**Table 1. Data pairs obtained from specific initial system setting $(S_1(t_1), E_1(t_1), I_1(t_1), R_1(t_1), \beta)$.**

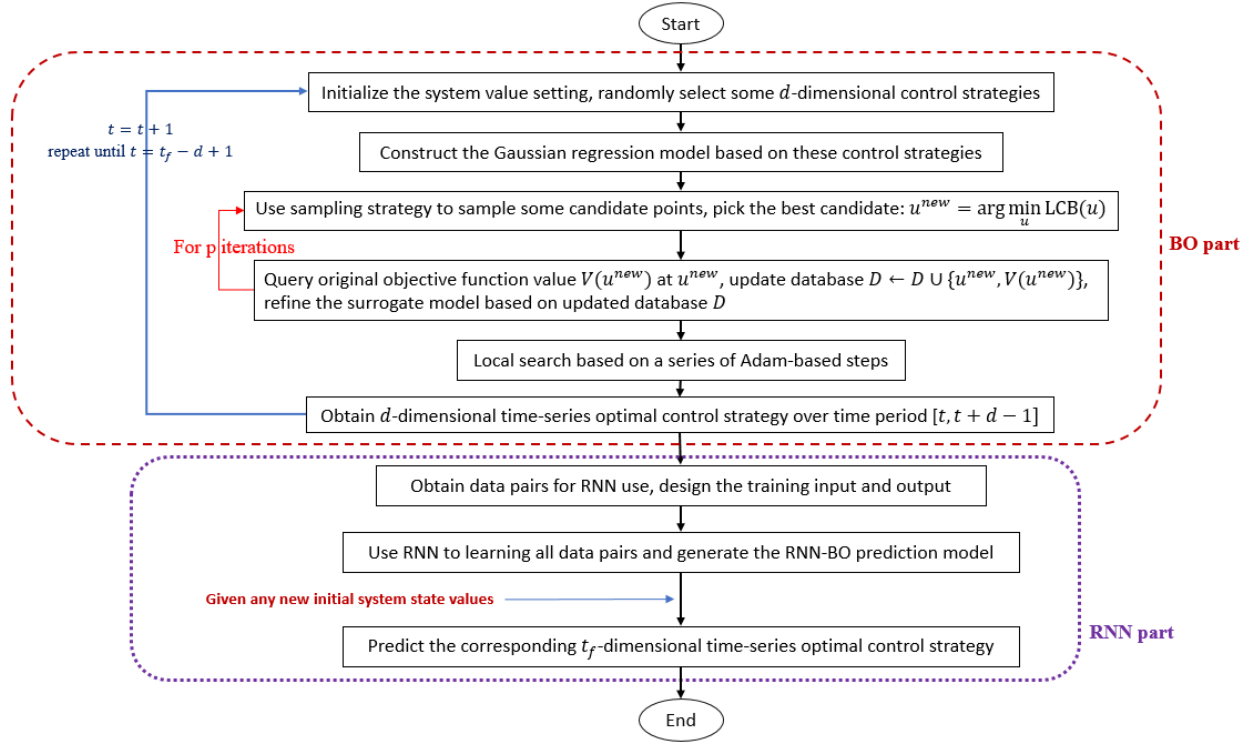

Fig. 3. Flowchart of the RNN-BO algorithm

## 4. Numerical Simulation

In this section, some simulation experiments are carried out to illustrate the efficiency and effectiveness of the RNN-BO algorithm. All experiments are implemented on Python version 3.7 with Intel Core i5 CPUs and 32G memory. The simulation experiments are carried out on the deterministic SEIR control model and the stochastic SIS control model. (Some extra simulation tests on some synthetic functions are also implemented, we will show the related results for reader's interests in **Appendix**.)

*4.1 Effectiveness of the RNN-BO algorithm*

This section tests the efficiency and effectiveness of the RNN-BO algorithm on the deterministic SEIR control model formulated in Eqns. (1) – (6). We also demonstrate that the optimal control generated by the BO algorithm for other initial system value settings may not be the optimal and effective for the model with new different initial system value setting. That's also the reason why we propose the RNN-BO algorithm, a model is capable to learn from historical data and predict for new situations.

Fig. 7 shows the trends of accumulated objective function values under different optimal control strategies. In the tests of this section, we set the system parameter infection rate as $\beta = 0.25$, $\beta = 0.3$, and $\beta = 0.4$. For each infection rate, vary the initial system state values. For example, in Fig. 7 (a), OptimalControl1 represents the optimal control strategy solved by the BO algorithm when the initial system state values are $S_1(t_1) = 0.4$, $E_1(t_1) = 0.13$, $I_1(t_1) = 0.47$, $R_1(t_1) = 0.0$, and system parameter $\beta = 0.25$. The line under OptimalControl1 means that the accumulated objective function values when we applied this OptimalControl1 to the model with new system state values $S(t_1) = 0.5$, $E(t_1) = 0.3$, $I(t_1) = 0.2$, $R(t_1) = 0.0$ and same system parameter $\beta = 0.25$. OptimalControl2 represents the optimal control strategy solved by the BO algorithm when the initial system state values $S_1(t_1) = 0.8$, $E_1(t_1) = 0.0$, $I_1(t_1) = 0.2$, $R_1(t_1) = 0.0$ and system parameter $\beta = 0.25$. The line under OptimalControl2 means that the accumulated objective function values when we applied this OptimalControl2 to the model with new system state values $S(t_1) = 0.5$, $E(t_1) = 0.3$, $I(t_1) = 0.2$, $R(t_1) = 0.0$ and same system parameter $\beta = 0.25$. OptimalControl3 is associated with the initial system state values $S_1(t_1) = 0.6$, $E_1(t_1) = 0.03$, $I_1(t_1) = 0.37$, $R_1(t_1) = 0.0$, OptimalControl4 is associated with initial system state values $S_1(t_1) = 0.3, E_1(t_1) = 0.3, I_1(t_1) = 0.4, R_1(t_1) = 0.0$, OptimalControl5 is associated with initial system state values $S_1(t_1) = 0.5, E_1(t_1) = 0.2, I_1(t_1) = 0.3, R_1(t_1) = 0.0$.

RNN-BO OptimalControl represents the optimal control strategy predicted by the RNN-BO prediction model, the line under RNN-BO OptimalControl means that the accumulated objective function values when we applied this RNN-BO OptimalControl to the model with new system state values $S(t_1) = 0.5, E(t_1) = 0.3, I(t_1) = 0.2, R(t_1) = 0.0$ and same system parameter $\beta = 0.25$. RealOptimalControl represents the actual optimal control strategy generated by the BO algorithm for the model with the new initial system state values $S(t_1) = 0.5, E(t_1) = 0.3, I(t_1) = 0.2, R(t_1) = 0.0$.

The small figures in Fig. 7 (a) – (c) are the zoom figures to show the trends clearly. From three figures (a) to (c), we can see that for the model with a new different initial system value setting, the optimal control strategies (OptimalControl1 – OptimalControl5) generated for other different initial system value settings don't perform good for the model with new initial system state value. This means that those optimal controls are the optimal control solutions for specific situation, they are not the optimal control solution for different situation. However, whatever the infection rate is, RNN-BO OptimalControl behaves similar effect as RealOptimalControl. That means although sometimes RNN-BO OptimalControl doesn't perform better than RealOptimalControl, the RNN-BO algorithm is flexible and accurate enough to predict the optimal solution or predict the solution closer to the actual optimal solution. Besides, different from other control strategies, the RNN-BO OptimalControl is predicted without going through the optimization iterations anymore, which is computationally efficiency.

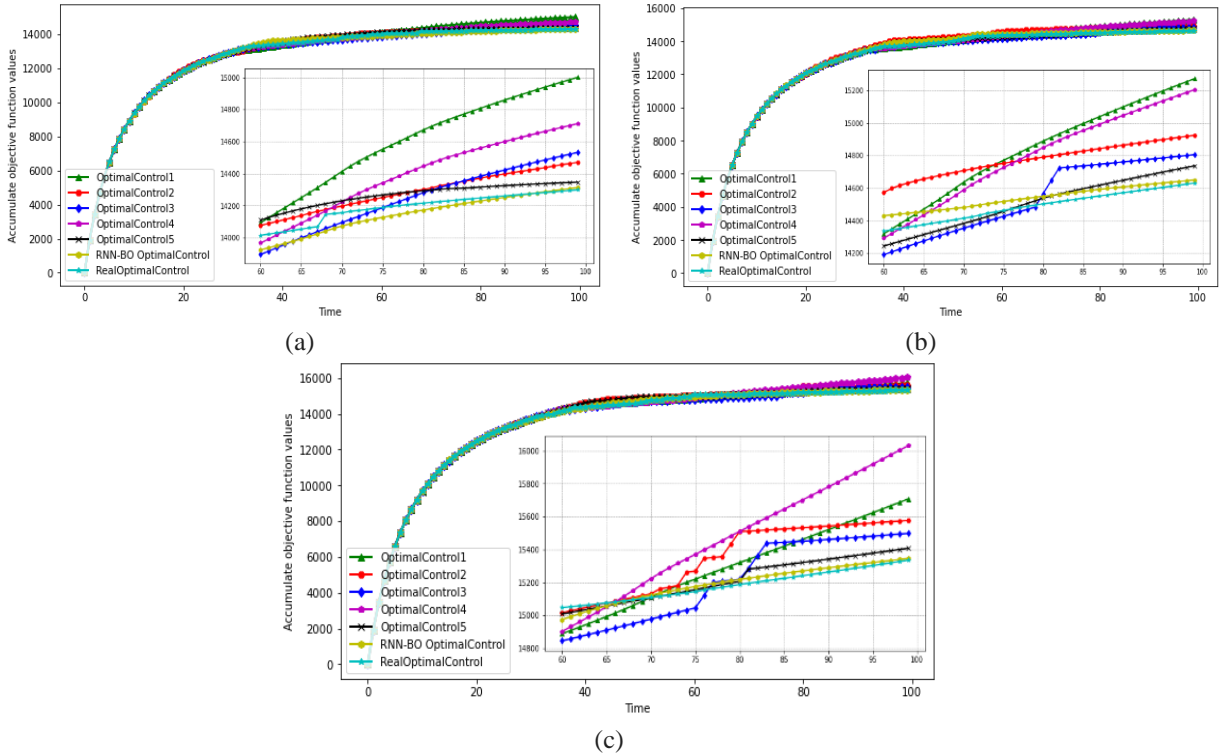

Fig. 7. (a) Accumulate objective function values generated by different optimal control when $\beta = 0.25$.
(b) Accumulate objective function values generated by different optimal control when $\beta = 0.3$.
(c) Accumulate objective function values generated by different optimal control when $\beta = 0.4$.

*4.2 Analysis of different RNN layers and different training epochs*
This section studies the impact of different RNN layers and training epochs on the algorithm's performances in deterministic SEIR control model. For all simulation tests in this section, the parameters of the RNN-BO prediction model are set as Dropout = 0.2, the activation = 'relu', the compile model optimizer = 'adam', loss = 'mse' during the implementation in Python. The meaning of those parameters is defined in [29]. Set the new initial system value setting need to be predicted is $S(t_1) = 0.5, E(t_1) = 0.3, I(t_1) = 0.2, R(t_1) = 0.0, \beta = 0.4$. Then use the prediction model with different RNN layers and different training epochs to predict the corresponding optimal control strategies, observe the training loss of the prediction model and the final best objective function values under these optimal control strategies.

First, we study the impact of different RNN layers on the training loss. In this simulation experiment, the training epochs is fixed as 9, the number of RNN layers that we test are 2, 3, 4, 5, 6, and 7. The simulation result is shown in Fig. 8 (a). The trends of the training loss of different RNN layers are almost similar. All of them decrease a lot during the first three training epochs then gradually subside after that. When the number of RNN layers is 7, the training loss is relatively worse than those in other situations.

Next, we study the impact of different training epochs on the final best objective function value. In this simulation experiment, the number of RNN layers of the prediction model that we test are 2, 3, 4, 5, 6, and 7. The training epoch that we test is from 1 to 16. The simulation result is shown in Fig. 8 (b). While the training epoch is 1, the best objective function values of different number of layers are all higher than 15500. However, when the number of layers is 3, 4, 5, and 6, if the training epoch is between 2 to 16, the RNN-BO algorithm archives robust performances, it always solves low final best objective function values. When the number of layers is 2 or 7, the results show that the RNN-BO algorithm doesn't perform well on the global optimization for the studied SEIR control model. The best objective function value is sensitive to the training epoch when the number of layers is 2 or 7.

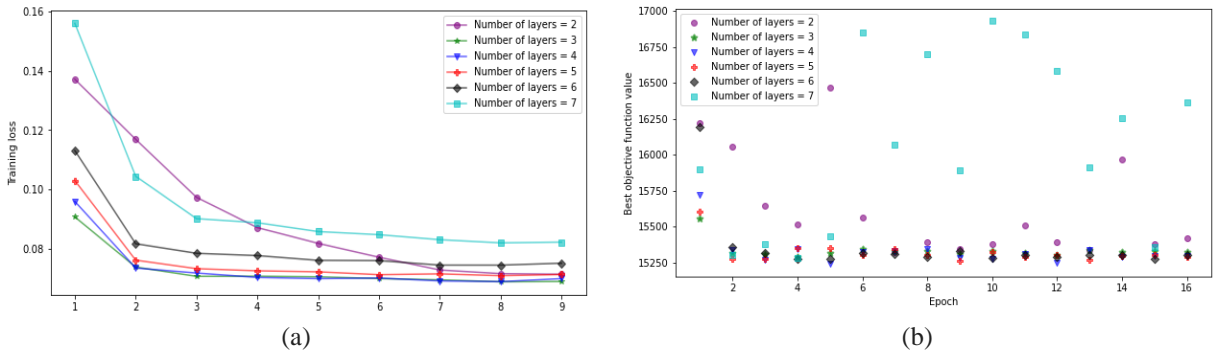

**Fig. 8. (a) Training loss of different RNN layers when the training epochs = 9.**
**(b) Best objective function value of different RNN layers under different number of training epochs.**

*4.3 Comparison with other algorithms*
In this section, we compare the RNN-BO algorithm to the standard Bayesian Optimization algorithm (standard BO algorithm) and a high-dimensional Bayesian Optimization algorithm proposed in [30]. For simplify, we name this high-dimensional Bayesian Optimization algorithm as 'Reference BO algorithm' in this section. Since we test the RNN-BO algorithm on the deterministic SEIR control model in section 4.1 and 4.2. To further prove the effectiveness of the RNN-BO algorithm on other model, we conduct the simulation of this section on the stochastic SIS control model formulated in Eqns. (10) – (13). All system parameter values remain unchanged for all tests in this section, the initial system values what we test is $S[0] = 0.6, I[0] = 0.4$. Referenced BO algorithm in its original paper is tested with different chosen dimensions $d$ [30]. Therefore, in here we also select different dimensions for the reference BO algorithm to do the comparison.

The simulation results are shown in Fig. 9. We can see that when the SIS model is without any control, the fraction of the infectious population over time behaves the oscillation property shown in Fig. 9 (a). It means that the epidemic would repeated outbreak. Also, the accumulated objective function value under null control condition keeps going up shown in Fig. 9 (c). When we use different Bayesian Optimization algorithms to solve the optimal control strategy, the corresponding fraction of the infectious population over time and the accumulated objective function value are obtained. For the Reference BO algorithm, e only show the results when the number of chosen dimensions $d$ is equal

to $5, 30, 60, 90$ out of $100$ dimensions. We tested $d = 5, 10, 20, 30, 40, 50, 60, 70, 80, 90$ for the Reference BO algorithm, it achieved the best objective function value when $d$ is equal to 30. From Fig. 9, we can see that all tested Bayesian Optimization algorithms significantly control the spread of epidemic. The corresponding accumulated objective function values over time also are about 6 times less than that when the model is without any control. For different Bayesian Optimization algorithms, the simulation results indicate that the Reference BO algorithm performs better than the standard BO algorithm, no matter what the number of chosen dimensions is. By comparing to the standard BO algorithm and the Reference BO algorithm, the RNN-BO algorithm achieves the best performance on controlling the spread of epidemic, decreasing infectious population, and minimizing the objective function value.

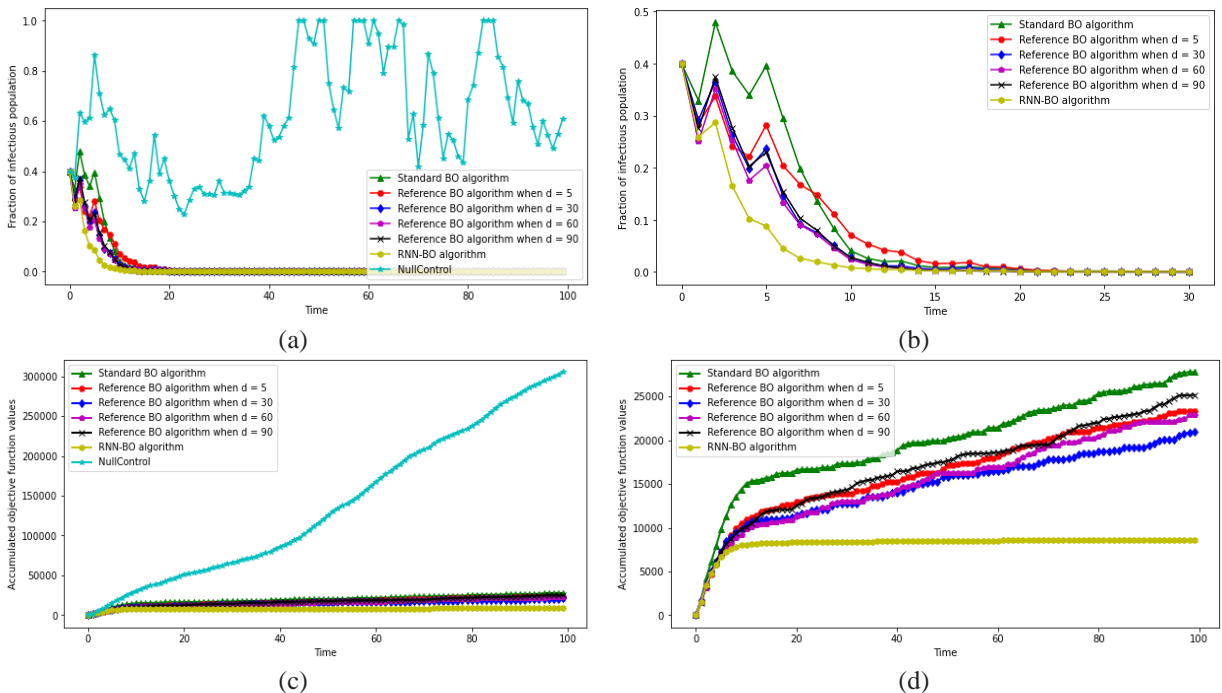

**Fig. 8. (a) The trends of infectious population over under different BO algorithms and null control condition.**
**(b) Zoom figure of the trends of infectious population under different BO algorithms.**
**(c) Accumulated objective function value under different BO algorithms and null control condition.**
**(d) Zoom figure of accumulated objective function value under different BO algorithms.**

## 5. Conclusions

In this paper, a new RNN-BO high-dimensional optimization algorithm is proposed by combining the high-dimensional Bayesian Optimization and recurrent neural network to improve computational efficiency and effectiveness. This proposed RNN-BO algorithm is flexible, it can be applied to predict the optimal control solution for different cities, counties, or countries. For example, the same epidemic outbreaks in different countries, the related disease data in the developed countries like the United States might be quickly available to be used, more adequate and comprehensive than some developing countries. Although the optimal control strategy plan of a specific place may not be the optimal or effective for other places, the RNN-BO algorithm can utilize the historical data from different places to develop a general and flexible RNN-BO prediction model. It means that for some developing countries, if they are not willing to spend a lot of time and money to collect the useful disease data, they can learn the available historical data of the same disease from other countries, and then generate the prediction model to predict the corresponding optimal control plan that is effective and applicable to their own countries. Some simulation experiments are implemented to prove that the RNN-BO algorithm is a promising approach. Different simulations also demonstrate the RNN-BO algorithm is robust and effective on the deterministic SEIR control model and stochastic SIS control model. We study the impact of different RNN layers and different training epochs on the RNN-BO algorithm's performances. In the future research, the RNN-BO algorithm should be studied and applied into more complicated and famous models. It would also be meaningful to design more effective sampling strategy, and tuning the parameters of the RNN-BO algorithm to speed up the calculations and predict the optimal result more accurate.

## References

1. Zhang, H., Yang, Z., Pawelek, K.A. and Liu, S., 2020. Optimal control strategies for a two-group epidemic model with vaccination-resource constraints. *Applied Mathematics and Computation*, *371*, p.124956.
2. Izzati, N., Andriani, A. and Robi'Aqolbi, R., 2020, October. Optimal control of diphtheria epidemic model with prevention and treatment. In *Journal of Physics: Conference Series* (Vol. 1663, No. 1, p. 012042). IOP Publishing.
3. Lü, X., Hui, H.W., Liu, F.F. and Bai, Y.L., 2021. Stability and optimal control strategies for a novel epidemic model of COVID-19. *Nonlinear Dynamics*, pp.1-17.
4. Wang, X., Peng, H., Shi, B., Jiang, D., Zhang, S. and Chen, B., 2019. Optimal vaccination strategy of a constrained time-varying SEIR epidemic model. *Communications in Nonlinear Science and Numerical Simulation*, *67*, pp.37-48.
5. Xu, D., Xu, X., Xie, Y. and Yang, C., 2017. Optimal control of an SIVRS epidemic spreading model with virus variation based on complex networks. *Communications in Nonlinear Science and Numerical Simulation*, *48*, pp.200-210.
6. Zugarini, A., Meloni, E., Betti, A., Panizza, A., Corneli, M. and Gori, M., 2020. An Optimal Control Approach to Learning in SIDARTHE Epidemic model. *arXiv preprint arXiv:2010.14878*.
7. Nabi, K.N., Kumar, P. and Erturk, V.S., 2021. Projections and fractional dynamics of COVID-19 with optimal control strategies. *Chaos, Solitons & Fractals*, *145*, p.110689.
8. Chen, Y., Bi, K., Wu, C.H.J., Ben-Arieh, D. and Sinha, A., 2021. High dimensional Bayesian Optimization Algorithm for Complex System in Time Series. *arXiv preprint arXiv:2108.02289*.
9. Moriconi, R., Deisenroth, M.P. and Kumar, K.S., 2020. High-dimensional Bayesian Optimization using low-dimensional feature spaces. *Machine Learning*, *109*(9), pp.1925-1943.
10. Zhang, M., Li, H. and Su, S., 2019. High dimensional Bayesian Optimization via supervised dimension reduction. *arXiv preprint arXiv:1907.08953*.
11. Li, C., Gupta, S., Rana, S., Nguyen, V., Venkatesh, S. and Shilton, A., 2018. High dimensional Bayesian Optimization using dropout. *arXiv preprint arXiv:1802.05400*.
12. Pandey, G., Chaudhary, P., Gupta, R. and Pal, S., 2020. SEIR and Regression Model based COVID-19 outbreak predictions in India. *arXiv preprint arXiv:2004.00958*.
13. Gray, A., Greenhalgh, D., Hu, L., Mao, X. and Pan, J., 2011. A stochastic differential equation SIS epidemic model. *SIAM Journal on Applied Mathematics*, *71*(3), pp.876-902.
14. https://www.cdc.gov/coronavirus/2019-ncov/hcp/infection-control-recommendations.html
15. Prem, K., Liu, Y., Russell, T.W., Kucharski, A.J., Eggo, R.M., Davies, N., Flasche, S., Clifford, S., Pearson, C.A., Munday, J.D. and Abbott, S., 2020. The effect of control strategies to reduce social mixing on outcomes of the COVID-19 epidemic in Wuhan, China: a modelling study. *The Lancet Public Health*, *5*(5), pp.e261-e270.
16. Shah, N.H., Suthar, A.H. and Jayswal, E.N., 2020. Control strategies to curtail transmission of covid-19. *International Journal of Mathematics and Mathematical Sciences*, *2020*.
17. Lemecha Obsu, L. and Feyissa Balcha, S., 2020. Optimal control strategies for the transmission risk of COVID-19. *Journal of biological dynamics*, *14*(1), pp.590-607.
18. Wang, X., Peng, H., Shi, B., Jiang, D., Zhang, S. and Chen, B., 2019. Optimal vaccination strategy of a constrained time-varying SEIR epidemic model. *Communications in Nonlinear Science and Numerical Simulation*, *67*, pp.37-48.
19. Lü, X., Hui, H.W., Liu, F.F. and Bai, Y.L., 2021. Stability and optimal control strategies for a novel epidemic model of COVID-19. *Nonlinear Dynamics*, pp.1-17.
20. Reluga, T.C., 2010. Game theory of social distancing in response to an epidemic. *PLoS computational biology*, *6*(5), p.e1000793.
21. Chen, Y., Bi, K., Wu, C.H.J., Ben-Arieh, D. and Sinha, A., 2021. A New Bayesian Optimization Algorithm for Complex High-Dimensional Disease Epidemic Systems. *arXiv preprint arXiv:2108.00062*.
22. Lee, D., Park, H. and Yoo, C.D., 2015. Face alignment using cascade gaussian process regression trees. In *Proceedings of the IEEE Conference on Computer Vision and Pattern Recognition* (pp. 4204-4212).
23. Moriconi, R., Kumar, K.S. and Deisenroth, M.P., 2020. High-dimensional Bayesian Optimization with projections using quantile Gaussian processes. *Optimization Letters*, *14*(1), pp.51-64.

24. Vakili, S., Liu, K. and Zhao, Q., 2013. Deterministic sequencing of exploration and exploitation for multi-armed bandit problems. *IEEE Journal of Selected Topics in Signal Processing*, *7*(5), pp.759-767.
25. Kingma, D.P. and Ba, J., 2014. Adam: A method for stochastic optimization. *arXiv preprint arXiv:1412.6980*.
26. Mikolov, T. and Zweig, G., 2012, December. Context dependent recurrent neural network language model. In *2012 IEEE Spoken Language Technology Workshop (SLT)* (pp. 234-239). IEEE.
27. Kamijo, K.I. and Tanigawa, T., 1990, June. Stock price pattern recognition-a recurrent neural network approach. In *1990 IJCNN international joint conference on neural networks* (pp. 215-221). IEEE.
28. Lipton, Z.C., Berkowitz, J. and Elkan, C., 2015. A critical review of recurrent neural networks for sequence learning. *arXiv preprint arXiv:1506.00019*.
29. https://www.tensorflow.org/api_docs/python/tf/keras/layers/SimpleRNN
30. Li, C., Gupta, S., Rana, S., Nguyen, V., Venkatesh, S. and Shilton, A., 2018. High dimensional Bayesian Optimization using dropout. *arXiv preprint arXiv:1802.05400*.

# Appendix

**Simulation of the RNN-BO algorithm on synthetic functions**

We test the solution accuracy generated by the RNN-BO algorithm on three high-dimensional synthetic functions (Rastrigin function, Rosenbrock function, Styblinski-Tang function). Due to the synthetic test functions don't contain constraints, there is no system state value designed in the input. We implement the tests by changing the initial start point of variable for each test function. Unless noted otherwise, we assume the variables of three test functions are 100 dimensions during simulation experiments. We conduct the simulation for each function across 5 runs to collect the data and 10 runs to evaluate the global optimal solution by changing the initial start points.

For Rastrigin function, the theoretical global optimal solution is $f(0,...,0) = 0$. For Rosenbrock function, the theoretical global optimal solution is $f(1,...,1) = 0$. For 100-dimensional Styblinski-Tang function, the theoretical global optimal solution is $f(-2.903534,...,-2.903534) = -3916.617$. During the historical data collecting process in BO part of the RNN-BO algorithm, we firstly randomly sample some points to construct the Gaussian process model, and then pick the best candidate at each iteration to optimize the acquisition function, finally do the local search to gradually converge to the global optimal point. To provide an intuitive display of the sampling point trajectory across one run in optimization part of the RNN-BO algorithm, we show a 3-dimensional plotting of each test function in Fig. 9. We can see that the RNN-BO algorithm always can figure out the global optimal point in finite iterations for each test function. For 10 evaluation runs to predict the global optimal solution by changing the initial start points, the results are summarized in Table 2.

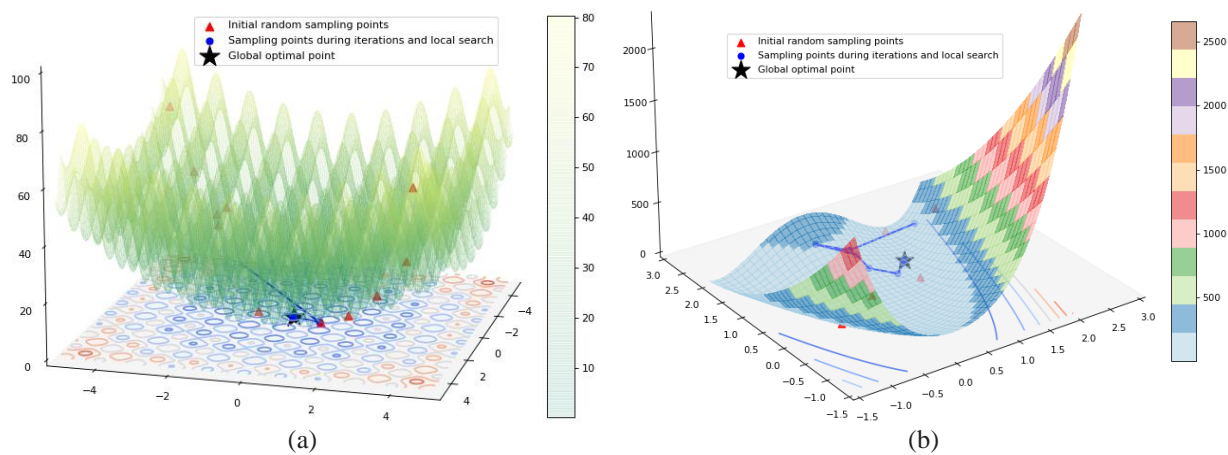

(a) (b)

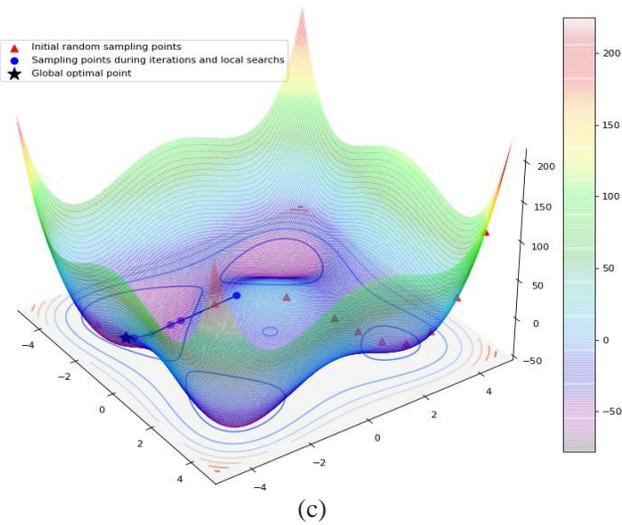
(c)

**Fig. 9. (a) 3-dimensional plotting of Rastrigin function.**
**(b) 3-dimensional plotting of Rosenbrock function.**
**(c) 3-dimensional plotting of Styblinski-Tang function.**

**Table 2. The optimal solutions of three synthetic functions crossing 10 runs**

| Function | Evaluation Dimension | Evaluation Domain | Evaluation Run | Evaluation Global Minimum | Plot Variable Dimension | Plot Domain |
|---|---|---|---|---|---|---|
| Rastrigin | 100 | [-5.12, 5.12] | 10 | $0.0 \pm 0.00003051$ | 2 | [-5.12, 5.12] |
| Rosenbrock | 100 | $[-\infty, \infty]$ | 10 | $0.000431 \pm 0.0013$ | 2 | [-1.5, 3] |
| Styblinski-Tang | 100 | [-5, 5] | 10 | $-3916.608 \pm 0.0007$ | 2 | [-5, 5] |